\documentclass{l4dc2025}

\title[Explaining RL with Alternatives]{``What are my options?'': Explaining RL Agents with Diverse Near-Optimal Alternatives (Extended)}
\usepackage{times}
\usepackage[noend]{algorithmic}
\usepackage{algorithm}
\usepackage{textcomp}
\usepackage{xcolor}
\usepackage{url}
\usepackage{dsfont}
\usepackage{amssymb}

\DeclareMathOperator*{\argmax}{arg\,max}

 \coltauthor{\Name{Noel Brindise} \Email{nbrindi2@illinois.edu}\\
  \Name{Vijeth Hebbar} \Email{vhebbar2@illinois.edu}\\
  \Name{Riya Shah} \Email{riyahs3@illinois.edu}\\
   \Name{Cedric Langbort} \Email{langbort@illinois.edu}\\
  \addr Dept. of Aerospace Engineering, University of Illinois Urbana-Champaign, Urbana, USA}

\begin{document}

\def\ddefloop#1{\ifx\ddefloop#1\else\ddef{#1}\expandafter\ddefloop\fi}

    \def\ddef#1{\expandafter\def\csname c#1\endcsname{\ensuremath{\mathcal{#1}}}}
    \ddefloop ABCDEFGHIJKLMNOPQRSTUVWXYZ\ddefloop

    \def\ddef#1{\expandafter\def\csname s#1\endcsname{\ensuremath{\mathsf{#1}}}}
    \ddefloop ABCDEFGHIJKLMNOPQRSTUVWXYZ\ddefloop

    \def\ddef#1{\expandafter\def\csname b#1\endcsname{\ensuremath{\mathbb{#1}}}}
    \ddefloop ABCDEFGHIJKLMNOPQRSTUVWXYZ\ddefloop

\maketitle
\thispagestyle{plain}

\begin{abstract}%
In this work, we provide an extended discussion of a new approach to explainable Reinforcement Learning called Diverse Near-Optimal Alternatives (DNA), first proposed at L4DC 2025. DNA seeks a set of reasonable ``options'' for trajectory-planning agents, optimizing policies to produce qualitatively diverse trajectories in Euclidean space. In the spirit of explainability, these distinct policies are used to ``explain'' an agent's options in terms of available trajectory shapes from which a human user may choose. In particular, DNA applies to value function-based policies on Markov decision processes where agents are limited to continuous trajectories. Here, we describe DNA, which uses reward shaping in local, modified Q-learning problems to solve for distinct policies with guaranteed $\epsilon$-optimality. We show that it successfully returns qualitatively different policies that constitute meaningfully different ``options'' in simulation, including a brief comparison to related approaches in the stochastic optimization field of Quality Diversity. Beyond the `explanatory' motivation, this work opens new possibilities for exploration and adaptive planning in RL.
\end{abstract}


\section{Introduction}
\par This paper provides extended discussion of a publication in the Learning for Dynamics and Control (L4DC) 2025, including proofs of key theorems and minor corrections (\cite{pmlr-v283-brindise25a}).
\par The field of explainable AI, which seeks to explain AI outputs and behavior to human users, is remarkably eclectic. Though commonly associated with the interpretation of neural networks, xAI encompasses many applications in explainable planning (XAIP) as well, where ``explanations'' describe plans, trajectories, or policies to characterize the intent or behavior of autonomous agents. In this work, we consider the common problem of RL agents on a Markov decision process (MDP). While these agents typically seek one optimal policy, we suggest that alternative policies may be of interest, particularly those with distinct behaviors but similar expected cost to some optimum: an \textit{explanation via alternatives}. 
\par Consider a route planning task for a ground vehicle. If the human operator is dissatisfied with a given plan (say it passes through risky terrain or an unwanted waypoint), the operator may want to assess potential alternatives before making a decision. Providing a user with alternatives also illuminates plan flexibility. A plan may be ``inflexible'' if it traverses states where there are limited choices of action, e.g. the vehicle is confined to a valley or moving along the edge of a cliff. This echoes the ``critical states'' proposed by Huang et al., where available actions have drastically different effects on cost (\cite{huang2018establishing}). In contrast, a plan may be called ``flexible'' if multiple reasonable policies are available, such as when different roads lead to the same destination in similar time. 
\par In this work, we pursue an explanatory method for \textbf{value-based} Reinforcement Learning agents which we call \textit{Diverse Near-Optimal Alternatives} (DNA). DNA aims to answer questions such as ``what policies/paths can I reasonably take from here?'' and ``how will the cost compare?'' Given an agent starts in a particular state, the method partitions the state space into distinct ``corridors,'' each of which will correspond to a possible policy option from that state. It then establishes ``local problems,'' adaptations of the original environment which reward the agent for successfully traversing the corridor. The resulting policies and the trajectories they produce are subject to a set of safety guarantees. 
\par In this paper, Section 2 gives an overview of the state of the art in Explainable Reinforcement Learning and Explainable AI Planning as it relates to our approach. Section 3 provides theoretical background, and Section 4 describes our proposed algorithm. A simple illustrative experiment is discussed in Section 5 in which a Q-learning agent is applied to a tabular environment. We also compare our method to related approaches from the field of Quality Diversity, which we briefly introduce in the next section.

\section{Previous Work}
Explainable Reinforcement Learning (XRL) and Explainable AI Planning (XAIP) aim to improve human understanding of autonomous system behavior. Though these fields have yet to identify unified goals or metrics for ``explainability,'' distinct branches have emerged (\cite{milani2022survey}, \cite{ijcai2020-669}, \cite{vouros2022explainable}). At the highest level, \textit{interpretable design} approaches (re)construct an agent to be inherently more explainable, while \textit{post-hoc} methods interpret the existing model. 
\par Post-hoc explanation alone encompasses a wide variety of approaches. It may describe agent behavior based on trajectory observations (\cite{brindise2023pointwise}, \cite{movin2023explaining}), ``highlight'' important moments (\cite{pierson2023comparing}), or seek causal relationships between variables (\cite{madumal2020explainable}). It may also assess the influence of trajectories on success probability (\cite{cruz2019memory}) or policy shape (\cite{deshmukh2022trajectory}). In many approaches, a human suggests an ``alternative'' policy or trajectory; the explanation then highlights infeasible segments (\cite{alsheeb2023towards}) or suggests environment changes to enable the suggestion (\cite{brandao2022not}, \cite{NEURIPS2022_dbef234b}). Reward shaping has also been used to drive an agent to desired waypoints (\cite{movin2023explaining}, \cite{8968488}). 

\par However, explanations which \textbf{seek and offer multiple policy suggestions} are rare in XAIP/XRL. This type of problem has been addressed e.g. in road navigation, where multiple route suggestions are commonly offered; unfortunately, the Dijkstra-based algorithms in such applications are limited to graph-traversal settings. For more general settings, a recent branch of stochastic optimization called Quality Diversity (QD) shows promise. QD solves for a set of high-performing (``quality'') policies which are behaviorally distinct (``diversity'') by iteratively populating a \textit{behavior} or \textit{feature} space with the best performers (\cite{mouret2015illuminating, chatzilygeroudis2021quality, 10.3389/frobt.2016.00040}). These methods suffer in the stochastic setting, however, as they heavily rely on a one-to-one policy-to-trajectory connection (\cite{flageat2023uncertain}). 

\par In contrast, our work is tailored to stochastic environments and addresses uncertain dynamics more directly than potential QD-inspired approaches. The local optimization problems defined by DNA encourage trajectories to remain within distinct corridors, leading to provable guarantees on trajectory shapes. Coupled with the proposed structure of the alternative policies, the reward-shaping scheme of DNA will also provide useful optimality guarantees. 

\section{Background}
 This work includes a simple proof of concept on a Markov decision process where trajectories are \textbf{Lipschitz continuous} with respect to the Manhattan norm (see \cite{asadi2018lipschitz}). In general, trajectories need only be continuous in a subset of the state space dimensions, i.e., diversity may be sought for the projections of trajectories onto a subset of spatial dimensions. In our application, value functions are estimated using Deep Q networks (DQN). 

\par N.B.: superscripts denote ``named'' constants and subscripts denote indices in a sequence, i.e. $(s_0, s_1, s_2) = (s^a, s^b, s^a)$ means that the second state in a sequence was State $s^b$.
\begin{definition}[Markov Decision Process]
    A Markov decision process (MDP) is a tuple $\mathcal{M}=\langle S, A, T, R\rangle$, where $S$ is a finite set of discrete states, $A$ is a finite set of actions, $T: S\times A \times S \rightarrow \mathbb{R}$ is a stochastic transition function, and $R: S\times A \rightarrow \mathbb{R}$ is a reward function.
\end{definition}

We will require that $R(\cdot,\cdot)\geq 0$. As our case study will simulate continuous trajectories on a grid, we specifically consider the common setting of gridworld-based RL:

\begin{definition}[MDP on a Grid]\label{MDPgrid}
    An MDP on a grid is an MDP where $S\subset \mathbb{Z}^K$ and the \textbf{neighbors} of any $s\in S$ are defined as $$\mathcal N (s) \triangleq \{s'\in S\ |\ \|s'-s\|_M = 1 \}$$ where $\|\cdot\|_M$ denotes the Manhattan distance. The transition function satisfies the property that, for any state $s\in S$, $T(s,a,s')>0$ for some $a\in A$ only if $s'\in \mathcal{N}(s)$.  
\end{definition}

For trajectories on the MDP on a grid, we can introduce a notion of grid-continuity:

\begin{definition}[Continuous on a Grid]
A trajectory $(s_{t},s_{t+1},...)$ is continuous on a grid if, for all $t$, 
$$s_{t+1}\in \mathcal{N}(s_t)$$
\end{definition}
This concept of continuity can be understood in terms of reachability on grid $\mathcal{M}$, where transitions are only possible between neighbors.


We will also require an optimal value function, necessitating additional definitions:
\begin{definition}[Optimal Q Function] 
    An optimal Q function $Q^*$ is a mapping of state-action pairs $Q : S \times A \rightarrow \mathbb{R}$ such that
    \begin{equation}\label{bellman}
        Q^*(s_t,a_t) = R(s_{t},a_t) + \gamma\mathbb{E}[ \max_{a'}Q^*(s_{t+1},a')].
    \end{equation}
for all $s\in S$, where $s_t\in S$ is the state at time $t$ and $\gamma\in [0,1]$ is a discount factor. On an MDP with transition function $T$,
\begin{equation}
    \mathbb{E}[ \max_{a'}Q^*(s_{t+1},a')] = \sum_{s'\in S} T(s_t, a_t, s') \max_{a'}Q^*(s',a')
\end{equation}

\end{definition}

\begin{definition}[Optimal Value Function]
    The optimal value function $V^*:S \to \bR$ is defined as 
    \begin{equation}
        V^*(s) = \max_{a\in A}Q^*(s,a).
    \end{equation}
\end{definition}
We define a \textbf{policy} as any mapping $\pi : S\rightarrow A$. An \textbf{optimal policy} $\pi^*$ must always take the action associated with the largest value of $Q^*$ at $s$, i.e. $\pi^*(s) = \argmax_{a\in A} Q^*(s, a)$ for all $s\in S$. Now, for any policy $\pi$, we may define a generic value function $V^\pi$:
\begin{definition}[Value Function for Policy $\pi$] \label{defn:value_pi}
    The value function $V^\pi:S \to \bR$ is defined for state $s_t\in S$ as 
    \begin{equation}
        V^\pi(s_t) = R(s_t,\pi(s_t)) + \gamma \sum_{s'\in S} T(s_t, \pi(s_t), s') V^\pi(s').
    \end{equation}
\end{definition}


We now move into the discussion of our proposed algorithm.

\section{Diverse Near-Optimal Alternative Policies via Corridor Search}
\subsection{Motivation: Diverse Trajectory Shapes through Corridors}
We suppose that a human user seeks \textbf{policy options} to create \textbf{distinct trajectory shapes} from an initial state. Often, trajectories in space are described using waypoints, e.g. in aviation; however, stochasticity means that any policy $\pi$ produces a \textbf{family} of trajectories, so a single policy cannot be associated with a waypoint sequence in a one-to-one manner. This will require some adaptation:
\begin{enumerate}
    \item \textbf{Waypoints $\rightarrow$ Way-regions:} rather than selecting states $s$ individually, we select larger \textit{way-regions} $W\subset S$ to describe and shape trajectories.
    \item \textbf{Sequences of Waypoints $\rightarrow$ Corridors:} we replace a sequence of waypoints with a \textit{corridor}, an object defined on subsets of $S$ using way-regions. We will alter $R$ to incentivize trajectories to remain inside a corridor and traverse from an initial state to an intermediate goal.
\end{enumerate}

\par Thus motivated, we will assess possible options via \textit{corridor search}: given a starting state $s_i$, we create corridors connecting $s_i$ to intermediate goals expressed as ``terminal'' way-regions $S_\Omega$. We then establish local problems using reward shaping. Training policies on these local problems results in a set of diverse, near-optimal options, providing the human an overview of the choices from $s_i$ and, by extension, the flexibility of the planning situation. 

\subsection{Definitions}

Given an agent on MDP $\mathcal{M}$ at state $s_i$, we seek alternative policies $\hat{\pi}$ which (i) have sufficiently high expected payoff from $s_i$ and (ii) produce diverse families of trajectories. First, taking $W$ as the set of all possible subsets of $S$, way-regions can be formalized via functions $b:S\rightarrow W$ which are defined for each corridor. Some regions are selected to be avoided $w\in W_{bad}$ and others as (intermediate) goal states $w\in W_{good}$. All $s\in S$ may then be organized into sets:
\begin{align}\notag
    S_{out} &= \{s\in w\ |\ w\in W_{bad}\}\\\label{eq:partition}
    S_{\Omega} &= \{s\in w\ |\ w\in W_{good} \}\\\notag
    S_{in} &= \{s\in S\ |\ s\not\in (W_{good}\cup W_{bad})\}\notag
\end{align}
Note that this forms a partition of $S$. Formally, a corridor is then:

\begin{definition}[Corridor]\label{corridorDef1}
Consider a subset of states $S_{in}\subset S$ with initial way-region $w_0\subset S_{in}$ and some $S_\Omega$ as defined in \eqref{eq:partition}. A corridor is the set of states $S_c\subseteq S$ such that $S_c =  S_{in}\cup S_\Omega$.
\end{definition}

\par \textit{Note on Selection of Corridors:} The selection of appropriate or useful $S_{in},S_\Omega$ is highly dependent on the specific application case and likely constitutes a separate study. For our case study, where grid-continuous trajectories are sought in 2D space, the partition of $S$ will be accomplished using square \textit{cells}, as an example.

\begin{definition}[Cell for Grid Case Study]
A cell centered at $s'$ is described by
\begin{equation}
    c(s') = \{s\in S\ |\ s'[k] - d \leq s[k] \leq s'[k] + d \quad \forall k\} 
\end{equation}
for selected distance $d$, where $s[k]$ is the coordinate of $s$ along the $k^{th}$ spatial dimension.
\end{definition}
For continuous trajectories, a sequence of cells is used to construct a \textit{continuous corridor}:

\begin{definition}[Continuous Corridor]\label{corridorDef} A continuous corridor of length $B$ is a corridor which is expressible as $S_c = \{s\in c\ |\ c\text{ in C}\}$, where $C = (c_0,...,c_B)$ is a sequence of cells in $\mathcal{C}$ and any consecutive cells $c_b, c_{b+1}$ in $C$ are adjacent. 
\end{definition}

Two cells $c_i,c_j$ are \textit{adjacent} if and only if there exist some $s^1\in c_i$, $s^2\in c_j$ such that $s^1, s^2$ are neighbors by Definition \ref{MDPgrid} and $c_i\neq c_j$.  Finally, at the end of a corridor, we place a terminal region $S_\Omega \subset S$. For our case study, we select one side of $c_B$, called a \textit{terminal edge}:

\begin{definition}[Terminal Edge for Grid Case Study]
For corridor $C= (c_0,...,c_B)$, a terminal edge $S_\Omega \subset S$ is the set of all states contained in an edge of $c_B$, i.e., for $c_B$ centered at $s'$,
\begin{equation}
    S_\Omega = \{s\in c_B\ |\ s[k]-s'[k] = \alpha d\}
\end{equation}
for a selection of $k$ and $\alpha \in \{-1, 1\}$. The set of all terminal edges for corridor $C$ is denoted $\mathcal{E}^C$ and is created by varying $k$.
\end{definition}

\par In summary, corridors in the grid case will be connected sets produced by placing a chain of square `cells' ($c\subset S$) through the state space. This is done such that $s_i$ is contained in the corridor and the edge of some final cell is chosen as goal region $S_\Omega$.

\par Now we consider the \textbf{cost} of potential policies. From a state $s_i$ and some optimal benchmark policy $\pi^*$, a policy $\pi$ is a reasonable choice only if it satisfies the criterion for \textit{$\epsilon$-optimality:}
\begin{definition}[$\epsilon$-Optimal Policy] A policy $\pi$ with corresponding value function $V^\pi$ is \textit{$\epsilon$-optimal} if for a given ${\epsilon\geq 0} $ it satisfies
\begin{equation}\label{offopt}
    \frac{V^\pi(s_i)}{V^*(s_i)} \geq  \epsilon.
\end{equation} 
\end{definition}
By this definition, any reasonable alternative must have an expectation which is sufficiently close to the benchmark optimum, determined by a user-defined $\epsilon$.

\subsection{Algorithm: Methodology and Guarantees}
 
\par We propose local $Q$ learning problems which incentivize policies to follow corridors. Informally, we seek policies which achieve a sufficient reward even when the system is altered such that (i) if the agent leaves the corridor at an $s\not\in S_\Omega$, $R(s)=0$ and the episode \textbf{immediately terminates}, and (ii) if the agent reaches $s\in S_\Omega$, it \textbf{receives reward $V^*(s)$} and terminates. Formally:

\begin{definition}[Local Problem] \label{def:localQ} For a corridor with a choice of $S_{in},S_\Omega$ as in \eqref{eq:partition}, the local problem is the problem solving for optimal local policy $\pi_L$ on MDP $\mathcal{M}_L$ with $S_L = S$, $A_L = A$,
    \[ T_L(s,a,s')=
        \begin{cases} 
            T(s,a,s') & s \in S_{in} \\ 
            \mathds{1}_{s'=s}  & \mathrm{otherwise,}
        \end{cases}
    \]
    
where $\mathds{1}_E$ is indicator over event $E$ and 
    \[ R_L(s,a)=
\begin{cases} 
      (1-\gamma) V^*(s) & s \in S_\Omega\\ 
      R(s,a) & s \in S_{in}\\
      0 & \mathrm{otherwise.}
   \end{cases}
\]

\end{definition}
\par (For the grid case study, interior states are $S_{in} = \{s\ |\ s\in (c\setminus S_\Omega), c\in C\}$ given a corridor with cell sequence $C=(c_0,...,c_B)$.)
\par Intuitively, the reward $(1-\gamma)V^*(s)$ is the infinite-trajectory equivalent to $V^*(s)$, as shown in the proof. We will refer to this particular reward as the ``happily-ever-after'' reward, a term which we will justify shortly.
\par Now, with $\pi_L$ defined, we can define a related policy on the global MDP:  
\begin{definition}[Alternative Policy]\label{altPol}
Consider auxiliary state $\Delta_t$ for trajectory $\rho = (s_0,...,s_t)$ and corridor with $S_{in},S_\Omega$, where $\Delta_0 = 0$ if $s_0\in S_{in}$ and $\Delta_0 = 1$ otherwise; and
\[\Delta_{t+1}=
\begin{cases} 
      1 & s_{t+1} \not\in S_{in} \\
      \Delta_t & \mathrm{otherwise.}
   \end{cases}
\]
Then an alternative policy $\hat{\pi}$ for the corridor, defined on augmented state $\lambda = (s[0],...,s[K],\Delta)^T$ takes the piecewise form
\[ \hat{\pi}
(\lambda) =
\begin{cases} 
      \pi_L(\cdot) & \Delta_t = 0 \\
      \pi^*(\cdot) & \mathrm{otherwise.}
   \end{cases}
\]
\end{definition}

Intuitively, an alternative policy from $s_i$ in a corridor follows a local policy ${\pi}_L$ until the trajectory exits the corridor or reaches the intermediate goal region, after which it follows a benchmark optimal policy $\pi^*$. Now, we recall that for an alternative policy $\hat{\pi}$ to be accepted, it must have comparable optimality to $\pi^*$ in line with \eqref{offopt}. This brings us to our first important guarantee.
\begin{figure}[t]
\centering
\includegraphics[width=.5\columnwidth]{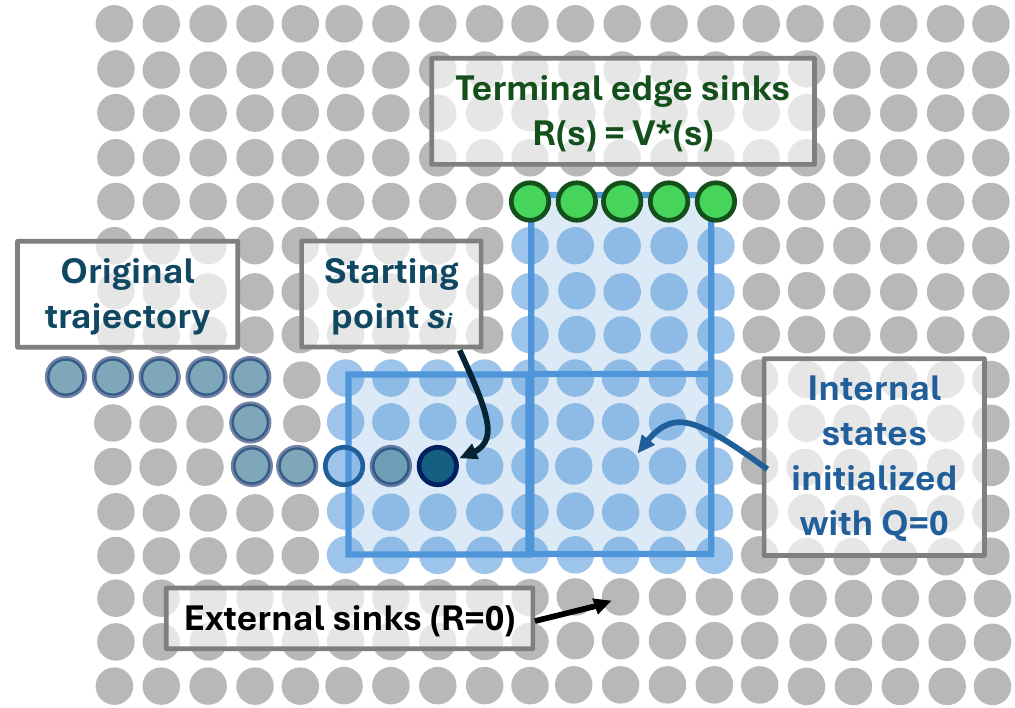}
\caption{\textbf{(Local Q-Learning Illustrated)} Simulated corridor $|C|=3$ with local MDP reward shaping applied.}
\label{localQ}
\end{figure}

\begin{theorem}[\textbf{$\epsilon$-Optimality Guarantee}]\label{thm:optLowerBoundAlt}
   Let $V_L^*$ be the value function corresponding to the local Q-learning problem in Definition \ref{def:localQ}. Then we have 
    $$V_L^*(s) \leq V^{\hat \pi}(( s, 0)^T)$$
    where $(s, \Delta)^T$ denotes $\lambda = ( s[0],...,s[K],\Delta)^T$ for $s\in S_\mathrm{in}$ given that $s$ has $K$ components and the inequality holds pointwise.

\end{theorem}
 
\par \textit{Theorem \ref{thm:optLowerBoundAlt}, informally:} Our method will train a local policy $\pi_L$ on a local problem in order to construct the final policy option $\hat{\pi}$. Theorem \ref{thm:optLowerBoundAlt} considers the expected reward $V^*_L$ found from optimizing $\pi_L$ \textbf{on the local problem} and compares it with the \textbf{(unknown) expected reward} $V^{\hat{\pi}}$ for the final policy. In particular, the theorem says that the performance of the full policy $\hat{\pi}$ will always be \textbf{at least as good} as the expectation found for the local policy. This is a direct consequence of our choice of $T_L$ and $R_L$ in Definition \ref{def:localQ}.

\par \textbf{Proof for Theorem \ref{thm:optLowerBoundAlt}}. The full proof is given in the Appendix. As a sketch, the theorem can be proven by defining and comparing several MDPs as follows:
\begin{itemize}
    \item $\cM$, the unaltered finite-horizon MDP of the original environment, where $\cM = \langle S,A,T,R,\gamma \rangle$
    \item $\cM_\lambda$, defined on state space $\Lambda$ of augmented states $\lambda = (s,\Delta)^T$. Importantly, the projection of $\cM_\lambda$ onto $S$ yields the same value function as $\cM$ from any starting state with $\Delta_0=0$. 
    \item $\Tilde{\cM_\lambda}$, where further alterations are made to the transition and reward functions. We show that the value functions for these $\cM_\lambda$ and  $\Tilde{\cM_\lambda}$ are identical from any starting state $\lambda_0=(s_0,0)^T$.
    \item $\Tilde{\cM}_{R_0}$, which is the same as $\Tilde{\cM_\lambda}$ except for a change to the reward function $\Tilde{R}_\lambda$. We show that $\Tilde{V}_{R_0}^\pi \leq \Tilde{V}_\lambda^\pi$ at all $\lambda$.
\end{itemize} 
\par The final theorem then gives $\Tilde{V}_{R_0}^\pi((s_0,0)^T) \leq {V^\pi}(s_0).$

\par We can also show that, given knowledge of the benchmark optimal policy and reward function, the probability that a trajectory remains within a corridor can be bounded from below:

\begin{theorem}[Bound on Success Probability of Trajectories in Corridors]\label{thm:safe}
For local value function $V^*_L$ on a corridor with terminal side $S_\Omega$, the probability $\mathbb{P}_{success}$ that a trajectory from $s_t$ reaches $S_\Omega$ is bounded by
    \begin{equation}
    \left(V^*_L(s_t)- \frac{\max(r_{in})}{1-\gamma}\right) \frac{1}{\gamma^\tau\max_{s\in S_\Omega}V^*(s)} \leq \mathbb{P}_{success}
\end{equation}
where $\max(r_{in}) = \max_{s\in S_{in}} R(s)$, $V^*$ is the value function for a benchmark optimal policy and $\gamma$ is the discount factor. $\tau\geq 0$ is any lower bound on the number of steps between $s_t$ and $s\in S_\Omega$.
\end{theorem}
Here, $\tau$ must be a lower bound for the shortest path from $s_t$ to $s\in S_\Omega$. Most conservatively $\tau\geq 0$; a first-order estimate for the grid is $\tau=\min_{s\in S_\Omega}||s_t-s||_M$. 

\par \textit{Theorem \ref{thm:safe}, informally:} Our previous theorem provided a guarantee on the expected reward given a starting state $s_i$. However, this is merely an \textbf{expectation}, and does not directly provide a guarantee of the frequency or probability with which the agent successfully traverses the corridor from $s_i$ to $S_\Omega$ (i.e., without first `falling out' into the region $S_{out}$). For instance, if an agent only reaches $S_\Omega$ in 1\% of cases but earns a very large reward when it does, training may obtain an acceptable $V^*_L$ despite undesirable or unexpected behavior. Theorem \ref{thm:safe} thus provides a bound on the probability of successful corridor traversal, which may be tightened when some lower bound on the number of steps between $s_i$ and $S_\Omega$ is known.

\subsection{Algorithm: Application and Complexity.}
\par Our algorithm seeks policies by searching over a set of corridors. For the cell-based example here, the number of corridors to check depends on corridor length and cell dimensions. In this demonstration, $S_c$ is created by uniformly placing square cells centered at distance $d-1$ from each other along each spatial dimension $k$; Figure \ref{localQ} shows this in $\mathbb{R}^2$. Each cell $c_b$ in a corridor has $2k$ possible terminal edges (and $2k$ possible $c_{b+1}$ to extend the corridor). Then, given starting state $s_i\in c_0$ and state space dimension $k$, there exist 
\begin{equation}\label{complexUpBound}
    n = \sum_{b=0}^B (2k)^{(b+1)}
\end{equation}
potential local problems. However, if the policy corresponding to $(c_0,...,c_\beta)$ is not $\epsilon-$optimal for any terminal edge of $c_\beta$ (Line \ref{lin:optimal-extension}), there can be no $\epsilon-$optimal policy for $(c_0,...,c_{\beta+1})$ by the optimality principle; thus, the search for $C = (c_0,...,c_{\beta},...)$ may be truncated, eliminating $\sum_{b=0}^{B-\beta} (2k)^{b+1}$ local problems. We also eliminate local problems where $c_0 \cup ... \cup c_B$ and $S_\Omega$ are identical to a previous problem (Line \ref{lin:avoid_backtrack}), since this represents the same local MDP and is thus redundant. In all, complexity will depend on the quantity of $\epsilon-$optimal options. Our example will have $k=2$, resulting in $n=\sum_{b=0}^B 4^b$ corridors to consider. 

\begin{algorithm}[t]
\caption{$\epsilon$-Optimal Alternative Policy Search}
\label{alg:explainable_Q}
\begin{algorithmic}[1]
\REQUIRE Environment $env$, $Q^*$, $\epsilon>0$, $s_0$, B, policy $\pi$.
\STATE Initialize $corridor\_stack = \big[(c_0)\big]$
\STATE Initialize $required\_corridors = \big[\big]$
\WHILE{$corridor\_stack$ is not empty}
    \STATE $curr\_corridor \gets$ pop first corridor from stack
    \IF{(length of $curr\_corridor$) $>B$ }
        \STATE \textbf{break}
    \ENDIF    
    \FOR{each $terminal\_edge$ of $curr\_corridor$} \label{lin:terminal_edge_explore}
        \IF{ $\max \{V^*(s)|s\in terminal\_edge\}<\epsilon V^*(s_0)$}
            \STATE \textbf{continue}
        \ENDIF
        \STATE Initialize $Q=0$, Set $Q(s,\cdot)=Q^*(s,\cdot) \text{ if } s\in terminal\_edge$ 
        \WHILE{$Q$ has not converged}
            \STATE \label{lin:init_Q_learn_episode} Initialize $s\in curr\_corridor$
            \WHILE{$s \in curr\_corridor$ and $s \not\in terminal\_edge$ } 
                \STATE Take step $s\to s'$ according to policy $\pi$.
                \IF{$s'\not\in curr\_corridor$} \STATE Set reward $0$ for step \ENDIF
                \STATE Perform Q-learning update
                \STATE Set $s=s'$
            \ENDWHILE
        \ENDWHILE
        \IF{$V(s_0)\geq\epsilon V^*(s_0)$} \label{lin:optimal-extension}
            \STATE Create cell $c'$ from $terminal\_edge$ away from $curr\_corridor$.
            \STATE Obtain $next\_corridor$ by appending $c'$ to $curr\_corridor$.
            \IF{$c'\in curr\_corridor$} \label{lin:avoid_backtrack}
                \STATE Ignore $terminal\_edge$ when exploring terminal edges (line \ref{lin:terminal_edge_explore}) of $next\_corridor$
            \ENDIF 
            \STATE Push $next\_corridor$ to $corridor\_stack$ 
            \IF{length of $next\_corridor=B$}
                \STATE Push $next\_corridor$ to $required\_corridor$
            \ENDIF 
        \ENDIF
    \ENDFOR
\ENDWHILE
\RETURN $required\_corridor$
\end{algorithmic}
\end{algorithm}

\section{Experimental Results}

\begin{figure}[t]
\centering
\begin{minipage}{0.45\textwidth}
        \centering
        \includegraphics[width=.8\textwidth]{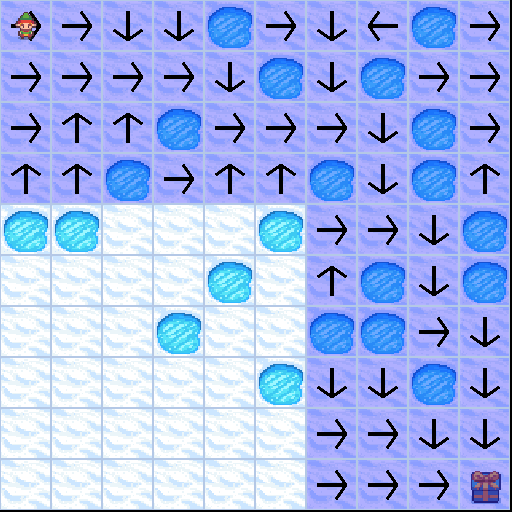}
\caption{\textbf{(Corridor 1 with $\epsilon=0.99$ sub-optimality.)} Arrows indicate the local policy actions.}
\label{fig:eps99}
    \end{minipage}\hfill
    \begin{minipage}{0.45\textwidth}
        \centering
        \includegraphics[width=.8\textwidth]
        {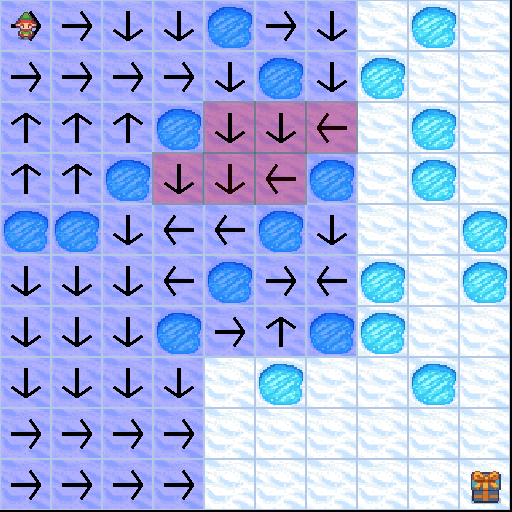}
\caption{\textbf{(Corridor 2 with $\epsilon=0.9$ sub-optimality.)} Differences from previous corridor are highlighted.}
\label{fig:eps90}
    \end{minipage}
\end{figure}

The results of Algorithm \ref{alg:explainable_Q} are now demonstrated for the Frozen Lake environment of OpenAI Gym. (See Github: \href{https://github.com/n-brindise/div-near-opt-alternatives}{n-brindise/div-near-opt-alternatives}, under construction as of 2025-06-11; contact nbrindi2@illinois.edu for help.) The agent begins at an \emph{initial state} in the top-left corner $(y,x) = (0,0)$ and attempts to reach the \emph{goal state} at the bottom-right corner $(9,9)$. \textit{Holes} in the frozen lake are absorbing states with reward 0. There are four actions $\{a_N, a_S, a_E, a_W\}$ corresponding to steps in cardinal directions $y-1, y+1, x+1, x-1$; the environment dynamics allow for ``slipping'' via a stochastic transition function. In this application, each action transitions a step in the intended direction with probability 0.9 and left or right of the intended direction each with 0.05. This example used tabular Q-learning; fundamentally, any method may be used which estimates a value function, such as a Deep Q-Network (DQN). 

\par Select experimental results for corridors with 5 cells ($B=4$) are highlighted in Figures \ref{fig:eps99} and \ref{fig:eps90}. Towards explanation, suppose a user asks what policy options exist from $(0,0)$. We first specify $\epsilon=0.99$, so that all solutions are required to have $V^*_L((0,0))$ very near the benchmark. This yields one option, shown in Figure \ref{fig:eps99}. Here, the corridor is highlighted in blue, with corresponding local policy $\pi_L$ shown by arrows; a left arrow indicates action $a_W$, up arrow $a_N$, and so on.
\par Now, supposing lower performance is acceptable, take $\epsilon=0.90$. This yields a second option (Figure \ref{fig:eps90}). This corridor terminates short of the goal (note that the agent can still \textit{reach} the goal upon exiting the corridor's terminal cell, corresponding to the piecewise policy switch from $\hat{\pi}=\pi_L$ to $\hat{\pi}=\pi^*$). The states highlighted red in Figure \ref{fig:eps90} have different optimal actions from Figure \ref{fig:eps99}, presumably resulting in more trajectories that proceed to the bottom left as opposed to the top right.

\par In the interest of safety, we may also assess how consistently any policy will follow a corridor to its terminus, i.e., the agent remains within $S_c$ until reaching $S_\Omega$. In our example, a user might prefer a more costly policy if it avoids the (dangerous) holes with higher probability as a matter of ``safety.'' \textbf{Theorem} \ref{thm:safe} yields the bounds in the \textbf{DNA} entry in Table \ref{tab:safety}. Note that $\tau$ for Corridor 1 is the Manhattan distance between $(0,0)$ and the nearest state in terminal edge $\{(9,6), (9,7), (9,8), (9,9)\}$. Similarly for Corridor 2, $\tau = 9$ is the Manhattan distance from $(0,0)$ to the edge $\{(6,3), (7,3), (8,3), (9,3)\}$. As reflected in the table, these bounds are indeed much lower than the experimentally-determined expectation for successful traversal.

\subsection{Comparison to Quality Diversity}
\par A basic Quality Diversity (QD) example was implemented for comparison with the DNA experimental setting. This was done using the Python QD toolbox pyribs (\cite{tjanaka2023pyribs}). QD methods rely on a descriptor function $\textbf{b} : U \rightarrow \mathcal{B}$, where $\mathcal{B}$ is a behavior space and $U$ is an observation of the trajectory space. In our case, $\mathbf{b}(\rho)$ identifies the corridor of length $i,\ 1\leq i \leq B$ corresponding to each trajectory. The function has the form $\mathbf{b} : \rho\rightarrow \{0,...,4\}^{B}$, and the process to map $\rho$ to $\mathcal{C}$ can be summarized as follows: (1) partition $S$ into cells by the same scheme as in DNA, (2) take $c_0 = c$ such that $s_0\in c$, (3) let $c' = c_0$, and (4) for each $s_t$ in $\rho = (s_0,...,s_t,...)$ and current $c'$: if $s_t\not\in c'$, set $c_{i}\leftarrow c$ s.t. $s_t\in c$ and set $c'\leftarrow c_i$. Increment $i$.
\par Given this mapping, the corridor descriptor is simply some $(b_1,...,b_B)$ where $b_i=0$ if cell $c_{i}$ does not exist, $b_i=1$ when cell $c_{i-1}$ is below $c_i$, $b_i=2$ when $c_{i-1}$ is right of $c_i$, and so on. Accordingly, a feature space can be defined on $\{0,...,4\}^{B}$, allowing for $5^{B-1}$ corridor configurations (when corridors are defined relatively from one origin point). Now, to capture policy costs, a fitness value $f_\theta$ is assigned as
\begin{equation}
    f_\theta(\rho) = \sum_{t=0} \gamma^t R(s_t)
\end{equation}
where $\rho = (s_0,...)$. Then the full QD problem seeks to solve
\begin{equation}
    \forall \mathbf{b} \in \mathcal{B}\quad \theta^* = \argmax_\theta f_\theta\quad\mathrm{s.t.}\quad \mathbf{b}=\mathbf{b}_\theta
\end{equation}
for each $\mathbf{b}$, where $\mathbf{b}$ is a point in $\mathcal{B}$. In this path-planning domain, the parameters $\theta$ correspond to control policy $\pi_\theta$. 

\par We briefly test whether QD identifies the corridors found by DNA, taking the Covariance Matrix Adaptation MAP-Annealing (CMA-MAE) variants proposed in \cite{tjanaka2023training}.  Results are shown in Table \ref{tab:safety} for separable CMA (CMA-Sep) over 3000 iterations and Limited Memory Matrix Adaptation (LM-MA) over 6000 iterations. LM-MA is the more successful of the two, identifying a policy which achieved 38\% consistency for Corridor 2. However, neither algorithm recovers reliable policies for Corridor 1. 

\begin{table}[h!]
\centering
\begin{tabular}{|c|c|c|}
\hline
 &\textbf{Corridor 1}&\textbf{Corridor 2}\\
 \hline
 \textbf{DNA}& 48.6\% (bound: 34.1\%) & 72.6\% (bound: 26.2\%)\\
 \hline
 \textbf{CMA-Sep} & 1.0\% & 2.0\%\\
  \hline
 \textbf{LM-MA} & - & 38.2\%\\
\hline
\end{tabular}
\caption{\label{tab:safety} Experimental probability that trajectory safely reaches $S_\Omega$ $(n=500)$}
\end{table}

\section{Conclusion and Future Work}
In this proof-of-concept example, our corridor search algorithm for continuous trajectories produced qualitatively distinct policies, successfully identifying ``alternative options'' from a state of interest on an MDP. The proposed local reward shaping problems satisfy a set of optimality and safety guarantees. Moreover, the method provides an interesting alternative to the evolutionary methods of Quality Diversity, optimizing local problems independently rather than via policy sampling and variation; this leads to more robust handling of stochasticity in experiment.
\par As this paper is conceptual in nature, future work is necessary to explore the applications of the method in experiment. A major candidate for future exploration is \textbf{improved exploration for Reinforcement Learning}; for instance, by establishing corridors in terms of abstract states corresponding to subtasks and/or task completion. 

\acks{This research was funded in part by a National Defense Science and
Engineering Graduate Fellowship and an Army Educational Outreach Program fellowship. The authors would also like to thank Andres Posada-Moreno for his helpful input.}

\bibliography{References}
\appendix
\section{Theorem \ref{thm:optLowerBoundAlt}}
We will prove the following:
   \par  Let $V_L^*$ be the value function corresponding to the local Q-learning problem in Definition \ref{def:localQ}. Then we have 
    $$V_L^*(s) \leq V^{\hat \pi}(( s, 0)^T)$$
    where $\lambda$ denotes $( s[0],...,s[K],\Delta)^T$ with $s\in S_{in}$ and the inequality holds pointwise.

\par \textbf{MDP 1: $\cM_\lambda$.} Consider the MDP $\cM = \langle S,A,T,R,\gamma \rangle$. We create a new MDP, $\cM_\lambda,$ by augmenting the state with an additional variable for ``switching'' purposes. This variable, $\Delta$, yields the full state $\lambda = (s, \Delta)^T$ (with the set of all augmented states notated $\Lambda$). 
\par We assign some subset $S_\Delta \subset S$. The variable $\Delta$ will track whether the agent has visited any $s\in S_\Delta$ via the function
\begin{flalign}
    \Delta' = f_\Delta(\lambda, s'), \quad f_\Delta(\lambda, s') := 
    \begin{cases}
        1 & s'\in S_\Delta\ \mathrm{or}\ \Delta = 1\\
        0 & \mathrm{otherwise}  
    \end{cases}
    \label{eq:delta}
\end{flalign}
where $s',\Delta'$ denote the values of $s,\Delta$ at the next time step. Intuitively, $\Delta=1$ if and only if the current state belongs to $S_\Delta$ or a previous state has visited it. 
\par \
\par \textbf{Aside: Existence of $t_\Delta$ and Uniqueness of $\rho_\lambda$.} Consider augmented trajectory $\rho_\lambda = (\lambda_0,\lambda_{1},...)$ with $\lambda_0=(s_0,\Delta_0)^T$ where $\Delta_0=0$. Firstly, we claim that if
\begin{equation}
    \lambda_{t+1}=(s_{t+1},\Delta_{t+1})^T, \quad \Delta_{t+1} = f_\Delta(\lambda_t,s_{t+1})
\end{equation}
for all $t\geq 0$, there exists some $t_\Delta$ such that, for all $t\geq 0$,
\begin{equation}\label{eq:monotonicDelta}
    \Delta_t=
    \begin{cases}
        0& t<t_\Delta\\
        1& \mathrm{otherwise}
    \end{cases}
\end{equation} 
or $\Delta_t=0$ everywhere. Secondly, we claim that, for a given $\rho=(s_0,s_1,...)$, there is one unique $\rho_\lambda$ which satisfies \eqref{eq:delta} on all $t$.
\par To prove the first claim, we will first show that $f_\Delta$ must be monotonically increasing over $\{0,1\}$, which is possible by induction.
\par \textbf{Show} $f_\Delta(\lambda_0,s_1)\leq f_\Delta(\lambda_1,s_2)$: Consider the following cases.
\begin{itemize}
    \item $s_1\not\in S_\Delta.$ By \eqref{eq:delta}, $f_\Delta((s_0,0)^T,s_1) = 0$. As $f_\Delta : \Lambda\times S\rightarrow \{0,1\}$, clearly $0 \leq f_\Delta(\lambda_1,s_2)$ regardless of $s_2$.
    \item $s_1\in S_\Delta.$ By \eqref{eq:delta}, $f_\Delta((s_0,0)^T,s_1) = 1$ and thus $\lambda_1=(s_1,1)^T$. Then $f_\Delta(\lambda_1,s_2)^T = 1$ and $f_\Delta((s_0,0)^T,s_1)\leq f_\Delta(\lambda_1,s_2)$.
\end{itemize}
\par\textbf{Prove} $f_\Delta(\lambda_t,s_{t+1})\leq f_\Delta(\lambda_{t+1},s_{t+2})$:
\begin{itemize}
    \item $\Delta_t=0:$ Our result follows from the logic for $\lambda_0$ above, with indices $0,1,2$ replaced by $t,t+1,t+2$.
    \item $\Delta_t=1:$ From \eqref{eq:delta}, we see that 
\begin{align}
    &f_\Delta((s_t,1)^T,s_{t+1}) = 1
    \quad\Rightarrow\quad \lambda_{t+1} = (s_{t+1}, 1)^T\\
    &\Rightarrow f_\Delta(\lambda_{t+1},s_{t+2}) = 1
\end{align}
and thus $f_\Delta(\lambda_t,s_{t+1})\leq f_\Delta(\lambda_{t+1},s_{t+2})$.
\end{itemize}
Therefore, $f_\Delta(\lambda_t,s_{t+1})\leq f_\Delta(\lambda_{t+1},s_{t+2})$ for all $t\geq 0$ when $\Delta_0=0$. To show the existence of $t_\Delta$ defined above, simply select 
\begin{equation}\label{eq:mintdelta}
    t_\Delta = \min (\{t\ |\ \Delta_t=1\})
\end{equation}
If the set is not empty, we have the minimum $t'$ such that $\Delta_t=0$ for all $t<t'$ by definition; by monotonicity, $\Delta_t=1$ for all $t\geq t_\Delta$. If the set is empty, this means that there exists no $t$ such that $\Delta_t=1$. As $f_\Delta$ maps to $\{0,1\}$, all $\Delta_t$ must therefore equal $0$, satisfying \eqref{eq:monotonicDelta} and therefore our claim. (QED 1) 
\par The second claim follows from our first claim. Any $\rho_\lambda$ which is not everywhere 0 must admit a $t_\Delta$ as above (or violate monotonicity). Since $t$ strictly increases in any sequence, \eqref{eq:mintdelta} must have a unique solution. Therefore, any $\rho$ will have exactly one corresponding $\rho_\lambda$ satisfying \eqref{eq:delta} at all $t$. (QED 2)

\par \ 
\par We now return to the definition of $\cM_\lambda$. We enforce transitions according to $f_\Delta$ using the \textbf{augmented transition function}
\begin{flalign}
    T_\lambda(\lambda,a,\lambda') =  
    \begin{cases}
        T(s,a,s') & \Delta'=f_\Delta(\lambda,s')\\
        0 & \mathrm{otherwise}  
    \end{cases}
\label{eq:TforMDP1}.
\end{flalign}
The \textbf{augmented reward function} remains the same:
\begin{flalign}
    R_\lambda(\lambda,a) = R(s,a)  
\label{eq:RforMDP1}.
\end{flalign}
Now we will consider the value function for trajectories on $\cM_\lambda$, comparing it to $\cM$ in Lemma \ref{VofMequaltoMlambda}.

\begin{lemma}\label{VofMequaltoMlambda}
    Given $\cM$ and $\cM_\lambda$ as defined above, 
    \begin{equation}
        V^\pi(s_t) = V^\pi_{\lambda}((s_t,\Delta_t)^T)\quad \text{if } \Delta_t =0.
    \end{equation}
    for a policy $\pi : \Lambda\rightarrow A$.
\end{lemma}
\textbf{Proof of Lemma \ref{VofMequaltoMlambda}:}
  For an infinite horizon MDP with reward function $\cR$, states $s\in\cS$, and infinite trajectories $\rho=(s_t,s_{t+1},...)$, the generic value function $V^\pi: \cS \to \bR$ for policy $\pi : \cS \rightarrow A$ can be expressed as 
\begin{flalign}
    V^\pi(s_t) = \mathbb{E}_\pi [\sum_{k=t}^\infty \gamma^{k-t} \mathcal{R}(s_{k}, \pi(s_{k}))] \label{eq:valuedef1}.
\end{flalign}
Denote the probability of $\rho = (s^a,s^b...)$ under policy $\pi$ as $p_{\rho|\pi}$. Let $\rho^i[k]$ denote the $k^{th}$ state in the $i^{th}$ trajectory $\rho^i$ and let $a^i_k$ be the policy action $\pi(\rho^i[k])$. Then \eqref{eq:valuedef1} can be rewritten for $\cM=\langle S,A,T,R,\gamma\rangle$:
\begin{equation}\label{eq:Vfororiginal}
V^\pi(s_t) = p_{\rho^1|\pi}\sum_{k=t}^{\infty} \gamma^{k-t} R(\rho^1[k],a^1_k)
    + p_{\rho^2|\pi}\sum_{k=t}^{\infty} \gamma^{k-t} R(\rho^2[k],a^2_k)+....
\end{equation}
where
\begin{equation}
    p_{\rho|\pi} = \prod_{k=0}^{\infty} T(\rho[k],a_k, \rho[k+1]).
\end{equation}
 For $\cM_\lambda$, we consider augmented trajectories $\rho_\lambda$ of $\lambda_t\in\Lambda$; these are generated according to $T_\lambda$ as in \eqref{eq:TforMDP1}. For augmented trajectory $\rho_\lambda=((s^a, \Delta^a)^T,(s^b, \Delta^b)^T,...)$ corresponding to $\rho=(s^a,s^b,...)$, we then have
 \begin{equation}
       p_{\rho_\lambda|\pi} = \prod_{k=0}^{\infty} T(\rho[k],a_k, \rho[k+1])=p_{\rho|\pi}
 \end{equation}
 if $\rho_\lambda$ is the unique trajectory with $\Delta_0=0$ and \eqref{eq:delta} true everywhere, and $p_{\rho_\lambda|\pi}=0$ otherwise. Therefore, 
\begin{align}
    V^\pi_\lambda((s_t,0)^T) \label{eq:mdp1expectation}&= p_{\rho^1|\pi}\sum_{k=t}^{\infty} \gamma^{k-t} R(\rho^1[k],a_k^1)
    + p_{\rho^2|\pi}\sum_{k=t}^{\infty} \gamma^{k-t} R(\rho^2[k],a_k^2)+...
\end{align}
which is exactly the same as \eqref{eq:Vfororiginal}. (QED)
 
\par \ 
\par \textbf{MDP 2: $\Tilde{\cM}_\lambda$.} We will now construct an altered MDP in which the transition and reward functions behave differently. For the transition function,
\begin{align}
    \Tilde{T}_\lambda(\lambda,a,\lambda') =  
    \begin{cases}
        T_\lambda(\lambda,a,\lambda') & \Delta=0\\
        1 & \Delta = 1,\ s=s', \Delta'=f_\Delta(\lambda,s')\\
        0 & \mathrm{otherwise}  \label{eq:TforMDP2}
    \end{cases}
\end{align}
This new $\Tilde{T}_\lambda$ simply makes the set $S_\Delta$ into absorbing states. The new reward function is
\begin{flalign}
    \Tilde{R}_\lambda(\lambda,a) = 
    \begin{cases}
        R(s,a) & \Delta = 0\\
        (1-\gamma)V^\pi_\lambda(\lambda) & \mathrm{otherwise}
    \end{cases}  
\label{eq:RforMDP2}
\end{flalign}
This reward assigns a discounted value function of $\cM_\lambda$ as a reward upon reaching any absorbing state  $s\in S_\Delta$. 

\par We will finally consider the value function and prove \ref{Vareequal}. \textbf{For the remainder of the proof,} we will allow $\rho$ to refer to trajectories over the augmented state space, i.e. $\rho=(\lambda_0,\lambda_1,...)$.
\begin{lemma}\label{Vareequal}
    Given $\cM_\lambda$ and $\Tilde{\cM}_\lambda$ as defined above, 
    \begin{equation}
    V^\pi_\lambda(\lambda_0)=\Tilde{V}^\pi_\lambda(\lambda_0)
    \end{equation}
\end{lemma}

\par From \eqref{eq:valuedef1} we may write a function similar to \eqref{eq:mdp1expectation}:
\begin{align}
    \Tilde{V}_\lambda^\pi(\lambda_t) \label{eq:mdp2expectation}&= p_{\rho^1|\pi}\sum_{k=t}^{\infty} \gamma^{k-t} \Tilde{R}_\lambda(\rho^1[k],a_k^1)
    + p_{\rho^M|\pi}\sum_{k=t}^{\infty} \gamma^{k-t} \Tilde{R}_\lambda(\rho^2[k],a^2_k)+...
\end{align}

If a $t_\Delta$ exists as defined previously, we may divide any trajectory originating at some $(s,0)$ on $\cM_\lambda$ or $\Tilde{\cM}_\lambda$ into two parts: a ``prefix,'' where $t<t_\Delta$ and $\Delta = 0$; and a ``suffix,'' where $t\geq t_\Delta$ and $\Delta=1$. We then have
\begin{flalign}
    \rho =& ( (s_0,0)^T, ...,(s_{t_\Delta-1},0)^T,(s_{t_\Delta},1)^T,...).
\end{flalign}
 If no such $t_\Delta$ exists, the prefix extends to infinity and the suffix has length 0.
 \par \ 
\par \textbf{Reward for Prefixes:} We will first consider \textbf{all prefixes,} which have the form $(s_0,0)^T, (s_1,0)^T,...$. From \eqref{eq:RforMDP1}, the reward for each state $\lambda=(s,\Delta)$ in $\cM_\lambda$ given policy $\pi$ is simply
\begin{equation}
    R(s,\pi(\lambda)).
\end{equation}
 \par For $\Tilde{\cM}_\lambda$, the reward $\Tilde{R}_\lambda$ from \eqref{eq:RforMDP2} is dependent on $\Delta$. Given that $\Delta=0$ on the entire prefix,
\begin{equation}
    \Tilde{R}_\lambda= R_\lambda= R(s,\pi(\lambda)). 
\end{equation}

\par \textbf{Probability of Prefixes:} From \eqref{eq:TforMDP2}, the transition function for $\Tilde{\cM}_\lambda$ is identical to  $\cM_\lambda$ when $\Delta=0$. Therefore, the probability of a given trajectory $\rho$ on both $\cM_\lambda$ and $\Tilde{\cM}_\lambda$ is
\begin{align}
    p_{\rho|\pi} =& \prod_{k=0}^{t_\Delta-1} T_\lambda(\lambda_k,\pi(\lambda_k), \lambda_{k+1})
    *\prod_{k=t_\Delta}^{\infty} \mathcal{T}_{suf}(\lambda_k,\pi(\lambda_k), \lambda_{k+1})
\end{align}
if $t_\Delta$ exists, where $\mathcal{T}_{suf}$ remains unknown. When $t_\Delta$ does not exist,
\begin{align}
    p_{\rho|\pi} =& \prod_{k=0}^{\infty} T_\lambda(\lambda_k,\pi(\lambda_k), \lambda_{k+1}).
\end{align}

For shorthand, we may rewrite this
\begin{equation}\label{eq:splitProb}
p_{\rho|\pi}=p_{\rho|\pi}^{pre}*p_{\rho|\pi}^{suf}
\end{equation}
where $p_{\rho|\pi}^{suf}=1$ in the latter case.
\par \ 
\par \textbf{Reward for Suffixes:} We now move to the suffixes.
 From \eqref{eq:RforMDP1}, the reward for $\cM_\lambda$ given policy $\pi$ is once again 
\begin{equation}
    R(s,\pi(\lambda))
\end{equation}
for each $\lambda_t$, $t\geq t_\Delta$. For $\Tilde{\cM}_\lambda$, \eqref{eq:RforMDP2} when $\Delta_t=1$ gives
\begin{flalign}
    \Tilde{R}_\lambda = (1-\gamma)V_\lambda^\pi(\lambda_t) 
\label{eq:RforsuffixMDP2}
\end{flalign}

\par \textbf{Probabilities for Suffixes:} In general, we have
\begin{align}
    p^{suf}_{\rho|\pi}=\prod_{k=t_\Delta}^{\infty} \mathcal{T}_{suf}(\lambda_k,\pi(\lambda_k), \lambda_{k+1})\notag
\end{align}
For $\cM_\lambda$, we simply have $\mathcal{T} = T_\lambda$ from \eqref{eq:TforMDP1}. For $\Tilde{\cM}_\lambda$, we have $\Delta=1$; thus, from \eqref{eq:TforMDP2}, only transitions with $s_t=s_{t_\Delta}$ have nonzero probability. Then
\begin{align}\label{eq:goodsuffix}
p^{suf}_{\rho|\pi}=
    \mathds{1}(s_t,s_{t_\Delta})
\end{align}
where $\mathds{1}:S\times S\rightarrow \{0,1\}$ is an indicator function with $\mathds{1}(s_i,s_j)=1$ when $s_i=s_j$ and $0$ otherwise.
\par \textbf{Putting the trajectories together:} We finally have expressions for $p_{\rho|\pi}$ for both MDPs. For $\cM_\lambda$, we see that $\cT$ is identical for prefix and suffix, so
\begin{equation}\label{eq:pfordmdp1}
    p_{\rho|\pi}= \prod_{k=0}^\infty T_\lambda(\lambda_k,\pi(\lambda_k), \lambda_{k+1}).
\end{equation}
For $\Tilde{\cM}_\lambda$,
\begin{equation}
    \Tilde{p}_{\rho|\pi} = \prod_{k=0}^{t_\Delta-1} T_\lambda(\lambda_k,\pi(\lambda_k), \lambda_{k+1})\prod_{k=t_\Delta}^{\infty} \mathds{1}(s_k,s_{t_\Delta})
\end{equation}
when $t_\Delta$ exists and $\Tilde{p}_{\rho|\pi}=p_{\rho|\pi}$ (as in \eqref{eq:pfordmdp1}) otherwise. Now we recall the value functions for $\cM_\lambda$ and $\Tilde{\cM}_\lambda$, given that trajectories start with $\lambda_0=(s_0,0)^T$. On the prefix, we have shown that $R_\lambda((s,\Delta)^T,a)=\Tilde{R}_\lambda((s,\Delta)^T,a) = R(s,a)$. Thus, each term of $V^\pi_\lambda$ as in \eqref{eq:mdp1expectation} becomes
\begin{align}\label{eq:mdp1vterm}
    p_{\rho^i|\pi}\bigg(\sum_{k=0}^{\infty} \gamma^k R(\rho^i[k],a^i_k)\bigg).
\end{align}
Now we consider the summands of $\Tilde{V}^\pi_\lambda$. For $\Tilde{V}^\pi_\lambda$, any $\rho^i$ where no $t_\Delta$ exists clearly result in terms identical to \eqref{eq:mdp1vterm}. For all other $\rho^i$ with nonzero probability (see \eqref{eq:goodsuffix}),
\begin{align}
     p_{\rho^i|\pi}\bigg(\sum_{k=0}^{t_\Delta-1} \gamma^k R(\rho^i[k],a^i_k) + \sum_{k=t_\Delta}^{\infty} \gamma^k (1-\gamma)V^\pi_\lambda(\rho^i[t_\Delta])\bigg)
\end{align}
Of these, we may \textbf{group} the terms with identical prefixes. Clearly, identical prefixes will have the same value for $t_\Delta$; for each such set of $\rho^i$, we then have a set-specific constant
\begin{equation}
R^{pre}=\sum_{k=0}^{t_\Delta-1} \gamma^kR(\rho^i[k])
\end{equation}
and can write the sum of terms in the value function $\Tilde{V}^\pi_\lambda$ for all $\rho^i$ in the set as
\begin{align}
     p_{\rho^1|\pi}&\bigg(R^{pre} + \sum_{k=t_\Delta}^{\infty} \gamma^k (1-\gamma)V^\pi_\lambda(\rho^1[t_\Delta])\bigg)\ +
     p_{\rho^2|\pi}\bigg(R^{pre} + \sum_{k=t_\Delta}^{\infty} \gamma^k (1-\gamma)V^\pi_\lambda(\rho^2[t_\Delta])\bigg)+...
\end{align}
By \eqref{eq:splitProb}, we may split all $p_{\rho^i|\pi}$ into $p^{pre}*p^{suf}$. As all prefixes are identical in our grouping, $p^{pre}$ is the same for each term; $p^{suf}$ will vary. Thus, with some relaxing of notation, we have
\begin{align}
     p^{pre}p_1^{suf}&\bigg(R^{pre} + \sum_{k=t_\Delta}^{\infty} \gamma^k (1-\gamma)V^\pi_\lambda(\rho^1[t_\Delta])\bigg)\ +
     p^{pre}p_2^{suf}\bigg(R^{pre} + \sum_{k=t_\Delta}^{\infty} ...\bigg)+...
\end{align}
Distributing the $p^{suf}_i$, this becomes
\begin{align}     p^{pre}\bigg(R^{pre}\sum_{i=1}^{I}p^{suf}_i + \sum_{i=1}^{I}p^{suf}_{i}\sum_{k=t_\Delta}^{\infty} \gamma^k (1-\gamma)V^\pi_\lambda(\rho^i[t_\Delta])\bigg)
\end{align}
for a set with $I$ trajectories.
First, we note that the full set of $p^{suf}$ must cover every possible suffix from $\lambda_n$ and therefore
\begin{equation}
    \sum_{i=1}^{I}p^{suf}_i = 1
\end{equation}
Then, distributing $p^{pre}$,
\begin{align}
=&p^{pre}R^{pre} + \sum_{i=1}^{I}p^{pre}p^{suf}_{i}\sum_{k=t_\Delta}^{\infty} \gamma^k (1-\gamma)V^\pi_\lambda(\rho^i[t_\Delta])\\
=&p^{pre}R^{pre} + \label{eq:termofvhere}\sum_{i=1}^{I}p_{\rho^i|\pi}\sum_{k=t_\Delta}^{\infty} \gamma^k (1-\gamma)V^\pi_\lambda(\rho^i[t_\Delta])
\end{align}
This may be further simplified using the sum of an infinite geometric series. Then

\begin{align}
    \sum_{k=t_\Delta}^{\infty} &\gamma^k (1-\gamma)V^\pi_\lambda(\rho^i[t_\Delta]) = \sum_{k=t_\Delta}^{\infty} \gamma^k V^\pi_\lambda(\rho^i[t_\Delta])-
    \sum_{k=t_\Delta+1}^{\infty} \gamma^{k} V^\pi_\lambda(\rho^i[t_\Delta])\\
    &=\bigg(\frac{V^\pi_\lambda(\rho^i[t_\Delta])}{1-\gamma}-\sum_{k=0}^{t_\Delta-1} \gamma^k V^\pi_\lambda(\rho^i[t_\Delta])\bigg) - \bigg(\frac{V^\pi_\lambda(\rho^i[t_\Delta])}{1-\gamma}-\sum_{k=0}^{t_\Delta} \gamma^k V^\pi_\lambda(\rho^i[t_\Delta])\bigg)\notag\\
    &=-\sum_{k=0}^{t_\Delta-1} \gamma^k V^\pi_\lambda(\rho^i[t_\Delta]) +\sum_{k=0}^{t_\Delta} \gamma^k V^\pi_\lambda(\rho^i[t_\Delta])\notag\\ &=\gamma^{t_\Delta}V^\pi_\lambda(\rho^i[t_\Delta])
\end{align}
Then we can rewrite \eqref{eq:termofvhere}:

\begin{align}\label{eq:MtildeVterm}
p^{pre}R^{pre} + \gamma^{t_\Delta}\sum_{i=1}^{I}p_{\rho^i|\pi}V^\pi_\lambda(\rho^i[t_\Delta])
\end{align}

\par We have already remarked that, for all $\rho$ with zero-length suffix ($t_\Delta$ nonexistent), $\cM_\lambda$ and $\Tilde{\cM}_\lambda$ have identical terms in $V$. We must still compare these remaining terms of $\Tilde{\cM}_\lambda$ (as in \eqref{eq:MtildeVterm}) to the terms for corresponding $\rho$ in $\cM_\lambda$. We may split these terms for $\cM$ into the same like-prefixed sets and rewrite them as we did for $\Tilde{M}_\lambda$. Then, each set can be expressed by

\begin{align}    \sum_{i=1}^I p_{\rho^i|\pi}\bigg(\sum_{k=0}^{\infty} \gamma^k R(\rho^i[k],a^i_k)\bigg) = p^{pre}R^{pre} + \sum_{i=1}^{I}p_{\rho^i|\pi}\sum_{k=t_\Delta}^{\infty} \gamma^k R(\rho^i[k],a^i_k)
\end{align}

Recalling the definition of $V^\pi_\lambda$ \eqref{eq:mdp1expectation}, we notice something quite familiar:
\begin{align}
=&p^{pre}R^{pre} + \sum_{i=1}^{I}p^{pre}p^{suf}_{i}\bigg(\gamma^{t_\Delta}\sum_{k=t_\Delta}^{\infty} \gamma^{k-t_\Delta} R(\rho^i[k],a^i_k)\bigg)\\
=&p^{pre}R^{pre} + \gamma^{t_\Delta}\sum_{i=1}^{I}V^\pi_\lambda(\rho^i[t_\Delta]).
\end{align}

which is exactly the same as \eqref{eq:MtildeVterm}. Having now compared all like-prefixed groupings of summands for both $V^\pi_\lambda$ and $\Tilde{V}^\pi_\lambda$, we may conclude that
\begin{equation}
V^\pi_\lambda(\lambda_0)=\Tilde{V}^\pi_\lambda(\lambda_0)
\end{equation}
for $\lambda_0 = (s_0,0)^T$, proving Lemma \ref{Vareequal}. (QED)
\par \ 

\par \textbf{Final MDP: $\Tilde{\cM}_{R_0}$.} We will finally establish an MDP which is identical to $\Tilde{\cM}_{\lambda}$ except for a change in the reward function. We define $S_\Omega\subset S_\Delta$, where $S_\Omega$ corresponds to the ``terminal edges'' in our algorithm. The reward function becomes
\begin{flalign}\label{eq:Rformdp3}
    \Tilde{R}_{R_0}(\lambda,a) = 
    \begin{cases}
        R(s,a) & \Delta = 0\\
        (1-\gamma)V^\pi_\lambda(\lambda_t) & \Delta=1,\ s\in S_\Omega\\
        0 & \mathrm{otherwise}
    \end{cases}  
\end{flalign}
In words, sink states $s\in S_\Delta$ which are not also in $S_\Omega$ now receive reward $0$. This brings us to our final lemma.

\begin{lemma}\label{2ndequalto3rd}
    $\Tilde{V}_{R_0}^\pi(s_0,0) \leq \Tilde{V}_\lambda^\pi(s_0,0)$.
\end{lemma}
As neither reward nor transition function change for the case $\Delta=0$, terms of the value function $\Tilde{V}_{R_0}^\pi$ with zero-length suffix are identical to $\Tilde{V}_\lambda^\pi$. For the rest, the contribution of each individual trajectory is 
\begin{align}
p_{\rho^i|\pi}\sum_{k=t_\Delta}^{\infty} \gamma^k \Tilde{R}_{R_0}(\rho^i[k], a^i_k).
\end{align}

We may compare each such contribution to the corresponding contribution by the same trajectory in $\Tilde{V}_\lambda^\pi$:
\begin{align}\label{eq:comparethelast}
    \Tilde{V}^\pi_{R_0}:\quad&p_{\rho^i|\pi}(\gamma^{t_\Delta}\Tilde{R}_{R_0}(\rho^i[t_\Delta],a^i_{t_\Delta}) + \gamma^{t_\Delta+1}\Tilde{R}_{R_0}(\rho^i[t_\Delta+1],a^i_{t_\Delta+1})+...)\\
\Tilde{V}^\pi_\lambda:\quad&p_{\rho^i|\pi}(\gamma^{t_\Delta}\Tilde{R}_\lambda(\rho^i[t_\Delta],a^i_{t_\Delta})\ \ + \gamma^{t_\Delta+1}\Tilde{R}_\lambda(\rho^i[t_\Delta+1],a^i_{t_\Delta+1})+...)\notag
\end{align}
Examining $\Tilde{R}_{R_0}$ and $\Tilde{R}_\lambda$, we readily observe that for all $\lambda=(s,\Delta)^T$, the pointwise inequality 
\begin{equation}
    \Tilde{R}_{R_0}(\lambda,\cdot)\leq\Tilde{R}_\lambda(\lambda,\cdot)
\end{equation}
 holds. Therefore, for all pairs of corresponding terms as in \eqref{eq:comparethelast},
 \begin{equation}
     \gamma^{t}\Tilde{R}_{R_0}(\rho^i[t],a^i_{t})\leq \gamma^{t}\Tilde{R}_\lambda(\rho^i[t],a^i_{t})
 \end{equation}
and thus
\begin{align}
p_{\rho^i|\pi}\sum_{k=t_\Delta}^{\infty} \gamma^k \Tilde{R}_{R_0}(\rho^i[k], a^i_k) \leq p_{\rho^i|\pi}\sum_{k=t_\Delta}^{\infty} \gamma^k \Tilde{R}_\lambda(\rho^i[k], a^i_k)
\end{align}
for all such trajectories. Then the full summation for $\Tilde{V}^\pi_\lambda$ and $\Tilde{V}^\pi_{R_0}$ brings us to
\begin{equation}
\Tilde{V}_{R_0}^\pi(\lambda_0) \leq \Tilde{V}_\lambda^\pi(\lambda_0)
\end{equation}
for all $\lambda_0=(s_0,0)^T$, and the lemma holds. (QED) We may now prove our theorem.
\par \ 
\par 
\textbf{Theorem \ref{thm:optLowerBoundAlt}}
($\Tilde{V}_{R_0}^\pi((s_0,0)) \leq V^\pi(s_0)$)

From Lemmas \ref{VofMequaltoMlambda}, \ref{Vareequal}, and \ref{2ndequalto3rd}, we have that 
\begin{equation*}
    V^\pi(s_0) = V^\pi_\lambda(s_0,0)=\Tilde{V}^\pi_\lambda(s_0,0)\geq\Tilde{V}^\pi_{R_0}(s_0,0)
\end{equation*}

for all $s_0\in S$. Thus
\begin{equation*}
    V^\pi(s_0) \geq \Tilde{V}^\pi_{R_0}(s_0,0)
\end{equation*}
at all $s_0\in S$. (QED)

\section{Theorem \ref{thm:safe}} 
\par \textbf{Outline:} To prove the theorem, we examine the expected reward for two categories of trajectories $\rho$ over $S$:
\begin{itemize}
    \item $\rho$ which reach a terminal state $s\in S_\Omega$ without leaving the corridor (``successes'')
    \item $\rho$ which either leave the corridor before reaching $S_\Omega$ or remain inside the corridor forever (``failures'')
\end{itemize}
We will show that the probability of success $\mathbb{P}_{success}$ can be bounded from below by calculating the minimum necessary proportion of successful trajectories to achieve the expectation $V^*_L(s_t)$.

Let $\cM_L = \langle S,A,T_L,R_L,\gamma \rangle$ be our local MDP and let  $V^*: S \to \bR$ be value function corresponding to some policy $\pi^*:S\rightarrow A$. 
The generic value function $V^\pi: S \to \bR$ for any policy $\pi:S\rightarrow A$ is  
\begin{flalign}
    V^\pi(s_t) = \mathbb{E}_\pi [\sum_{k=t}^\infty \gamma^{k-t} R_L(s_{t+k}, \pi(s_{t+k}))] \label{eq:valuedef2}.
\end{flalign}
Now, for any trajectory $\rho$ originating at some $s_t$, let $\rho[k]$ denote state $s_{t+k}$ in the trajectory and $a_k=\pi(\rho[k])$. Denote the probability of $\rho$ occurring from $s_t$ under policy $\pi$ as $p_{\rho|\pi}$. Then \ref{eq:valuedef2} can be rewritten as a sum over the possible trajectories from $s_t$, as follows:
\begin{flalign}
    V^\pi(s_t) = p_{\rho_1|\pi}\sum_{k=t}^\infty \gamma^{k-t} R_L(\rho_1[k],a_k^1) + p_{\rho_2|\pi}\sum_{k=t}^\infty \gamma^{k-t} R_L(\rho_2[k],a^2_k) +...\label{eq:expectationnew}
\end{flalign} 
\par We are interested in the \textbf{probability that our trajectories remain within the corridor}. Thus, we separate these terms into two groups: one in which all trajectories terminate at the desired terminal edge (\textit{success}), and one in which they either remain in the corridor forever or exit at a non-terminal boundary (\textit{failure}). We may then group $\rho$ such that all $\rho^i\in I$ are \textit{successes} and all $\rho^j\in J$ are \textit{failures}. Then
\begin{flalign}
   \mathbb{P}_{success}=\sum_{\rho^i\in I}^\infty p_{\rho^i|\pi}\quad\mathrm{and}\quad\mathbb{P}_{fail}=\sum_{\rho^j\in J}^\infty p_{\rho^j|\pi}
\end{flalign}
We will now examine the trajectories in each case. Let $s_{in}\in S_{in}$ denote any (non-terminal) state inside of the corridor and $s_\Omega\in S_\Omega$ denote any state in the terminal edge. Finally, $s_{out}\in S_{out}$ will be any state outside of the corridor, such that $S_{out} = S \setminus (S_{in}\cup S_\Omega$). Given $\rho=(s_0,s_1,...)$ is a \textbf{success,} there must exist some 
\begin{equation}\label{eq:tomegamin}
    t_\Omega = \min(\{t\ |\ s_{t}\in S_\Omega\}).
\end{equation}
Denote the particular $s$ reached at $t_\Omega$ by $s_\Omega$. From the reward function $R_L$, we then have
\begin{flalign}
    \sum_{k=0}^{\infty} \gamma^k R_L(\rho[k],a_k) = \sum_{k=0}^{t_\Omega-1} \gamma^k R(s_{in},a_k) + \sum_{k=t_\Omega}^{\infty} \gamma^k (1-\gamma)V^*(\rho[k]).
\end{flalign}
In words, all states preceding the terminal state must belong to $S_{in}$; otherwise, \eqref{eq:tomegamin} would not hold (or $\rho$ would not be a success). Now, we have by the transition function that all $s\in S_\Omega$ are absorbing, and thus $\rho[k]=\rho[t_\Omega] = s_\Omega$ for all $k\geq t_\Omega$. We see then that the last term can be simplified using the convergence of geometric series:
\begin{flalign}
    \sum_{k=0}^{\infty} \gamma^k R_L(\rho[k],a_k) = \bigg(\sum_{k=0}^{t_\Omega-1} \gamma^k R(s_{in},a_k)\bigg) + \gamma^{t_\Omega} V^*(s_\Omega).
\end{flalign}
 Letting $\max(r_{in}) := \max_{s\in S_{in}}R(s,\pi(s))$ and $\max(r_\Omega) := \max_{s\in S_\Omega}V^*(s)$, we can establish an upper bound on the discounted reward for any successful trajectory:
\begin{flalign}
    \sum_{k=0}^{\infty} \gamma^k R(\rho[k],a_k) \leq \sum_{k=0}^{t_\Omega-1} \gamma^k \max(r_{in}) + \gamma^{t_\Omega} \max(r_\Omega).
\end{flalign}
Moreoever, the value $\max(r_{in})$ is constrained to be a nonnegative constant and $0\leq \gamma < 1$, meaning that
\begin{flalign}
    \sum_{k=0}^{t-1} \gamma^k \max(r_{in}) \leq \sum_{k=0}^{t} \gamma^k \max(r_{in})
\end{flalign} 
for all $t$. Again by geometric series,
\begin{flalign}
    \sum_{k=0}^{t_\Omega-1} \gamma^k \max(r_{in}) < \frac{\max(r_{in})}{1-\gamma}\label{eq:geometric}
\end{flalign}
for any $t_\Omega$.
\par Since $t_\Omega$ may be as small as $0$, the term $\gamma^{t_\Omega} \max(r_\Omega)$ is not as easy to bound without additional knowledge of the dimensions of the corridor and $t_\Omega$. If a shortest path is known such that the distance between $s_t$ and $S_\Omega$ cannot be traversed in less than $d$ steps, 
\begin{flalign}
    \gamma^{t_\Omega} \max(r_\Omega) \leq \gamma^d\max(r_\Omega)
\end{flalign}
where $d=0$ if no such minimum step count is known. Applying these bounds to the full set of successful trajectories yields

\begin{flalign}
    p_{\rho_1|\pi}\sum_{k=0}^{\infty} \gamma^k R_L(\rho_1[k],a_k^1) + p_{\rho_2|\pi}\sum_{k=0}^{\infty} \gamma^k R_L(\rho_2[k],a_k^2)+...\\
    \leq \mathbb{P}_{success}(\frac{\max(r_{in})}{1-\gamma} + \gamma^d\max(r_\Omega)).\label{eq:boundonsucc}
\end{flalign}
Now we consider \textbf{``failing'' trajectories}. These may remain within $S_{in}$ forever, i.e.
\begin{flalign}
    \sum_{k=0}^{\infty} \gamma^k R_L(\rho[k],a_k) = \sum_{k=0}^{\infty} \gamma^k R(s_{in},a_k),\label{eq:inforever}
\end{flalign}
In this case, the reward may be bounded with the same geometric series:
\begin{flalign}
    \sum_{k=0}^{\infty} \gamma^k R_L(\rho[k],a_k) \leq \frac{\max(r_{in})}{1-\gamma} 
\end{flalign}

Alternatively, failed trajectories may exit the corridor at some non-terminal state, i.e.
\begin{flalign}
    \sum_{k=0}^{\infty} \gamma^k R_L(\rho[k],a_k) = \sum_{k=0}^{T-1} \gamma^k R(s_{in},a_k) + \sum_{k=T}^\infty\gamma^k R(s_{out},a_k).\label{eq:fallsout}
\end{flalign}
given the first $s\not\in S_{in}$ occurs at time $T$. In this case, the reward may be once again bounded similarly to successful trajectories, using $s_{out}\in S_{out}$ absorbing but noting this time that $R_L(s_{out},\cdot)=0$. Then
\begin{flalign}
    \sum_{k=0}^{\infty} \gamma^k R_L(\rho[k],a_k) \leq \frac{\max(r_{in})}{1-\gamma} + 0
\end{flalign}
Thus, returning to \ref{eq:expectationnew} and noting that $\mathbb{P}_{fail} = 1-\mathbb{P}_{success}$, we have
\begin{flalign}
    V^\pi(s_t) \leq \mathbb{P}_{success}(\frac{\max(r_{in})}{1-\gamma} + \gamma^d\max(r_\Omega)) \\
    + (1-\mathbb{P}_{success})(\frac{\max(r_{in})}{1-\gamma})\notag
\end{flalign}
Simplifying,
\begin{flalign}
    V^\pi(s_t) \leq \mathbb{P}_{success}(\gamma^d\max(r_\Omega))
    + \frac{\max(r_{in})}{1-\gamma}
\end{flalign}
As given in the local problem, we may replace $\max(r_\Omega)$ with $\max_{s\in S_\Omega}V^*(s)$, where $V^*$ is the optimal value function for the global policy on the original MDP. This yields the bound
\begin{flalign}
    (V^\pi(s_t)- \frac{\max(r_{in})}{1-\gamma}) \frac{1}{\gamma^d\max_{s\in S_\Omega}V^*(s)} \leq \mathbb{P}_{success}
\end{flalign}
In our specific case where $\max(r_{in}) \leq 1$, this becomes
\begin{flalign}
    (V^\pi(s_t)- \frac{1}{1-\gamma}) \frac{1}{\gamma^d\max_{s\in S_\Omega}V^*(s)} \leq \mathbb{P}_{success}
\end{flalign}
Thus, for any corridor, we have established a lower bound on the probability that a trajectory successfully remains within the corridor until reaching the terminal edge. (QED)

\end{document}